\pdfoutput=1

\documentclass[11pt]{article}

\usepackage[]{ACL2023}

\usepackage{times}
\usepackage{latexsym}

\usepackage[T1]{fontenc}

\usepackage[utf8]{inputenc}

\usepackage{microtype}

\usepackage{inconsolata}
\usepackage{graphicx}
\usepackage{tabularx}
\usepackage{subfigure}
\usepackage{stfloats}
\usepackage{booktabs}
\usepackage{amsmath,amsfonts,amssymb}

\title{Never Lost in the Middle: Mastering Long-Context Question Answering with Position-Agnostic Decompositional Training}

\author{Junqing He,\hspace{1em} Kunhao Pan, \hspace{1em}Xiaoqun Dong, \hspace{1em} \\ {\bf Zhuoyang Song,\hspace{1em} } 
{\bf Yibo Liu,\hspace{1em}} 
    {\bf Qianguo Sun,\hspace{1em}} \\
{\bf Yuxin Liang,\hspace{1em}} {\bf Hao Wang,\hspace{1em} } {\bf Enming Zhang,\hspace{1em}} {\bf Jiaxing Zhang} 
\\
        International Digital Economy Academy, Shenzhen, China\\
        \text\tt hejunqing@idea.edu.cn}

\begin{document}
\maketitle
\begin{abstract}
While large language models (LLMs) are equipped with longer text input capabilities than before, they struggle to seek correct information in long contexts. The "lost in the middle" problem challenges most LLMs, referring to the dramatic decline in accuracy when correct information is located in the middle. To overcome this crucial issue, this paper proposes to enhance the information searching and reflection ability of LLMs in long contexts via specially designed tasks called \underline{\textbf{P}}osition-\underline{\textbf{A}}gnostic \underline{\textbf{M}}ulti-step QA (PAM QA). Trained with this task, our model excels in focusing more precisely on the desired information. Experimental results show substantial improvement in Multi-doc QA and other benchmarks, surpassing state-of-the-art models by a 13.7\% absolute gain in shuffled settings and by 21.5\% in the passage retrieval task. We release our model and code to promote related research in the community.\footnote{ It is publicly available at  \url{https://huggingface.co/IDEA-CCNL/Ziya-Reader-13B-v1.0} and \url{https://github.com/hejunqing/never-lost-in-the-middle}}  
\end{abstract}

\section{Introduction}

Large Language Models (LLMs), renowned for their exceptional generative and zero-shot learning abilities across diverse natural language processing (NLP) fields, have found extensive downstream applications \cite{openai_2023, Boiko_MacKnight_Gomes_2023,cheng2023exploring,waisberg2023gpt,hu2023chatdb}.
However, LLMs suffer from severe hallucinations, significantly compromising their performance in knowledge-oriented QA, dialogue, and writing \cite{roberts2020much,agrawal2023language}. Retrieval Augmented Generation (RAG) is an effective solution to hallucinations, and remarkable improvements have been achieved by incorporating supporting knowledge into the input of LLMs \cite{Lewis_Perez_Piktus_Petroni_Karpukhin_Goyal_Küttler_Lewis_Yih_Rocktäschel_etal._2020, shuster2021retrieval,thoppilan2022lamda,Shi_Han_Lewis_Tsvetkov_Zettlemoyer_Yih_2023}. The most fundamental challenge to address in RAG is long context and Multi-document question answering (Multi-doc QA). 

Some research works around the problem with a complicated pipeline or system \cite{chen2023walking,lee2024humaninspired}, but we aim to improve foundation models as they are a core component of those methods. Thorough research has been conducted to deal with long context inputs, categorized into three mainstreams: The first is to expand the context window using a sliding window \cite{dai2019transformer,xiao2023efficient}. Other researchers proposed to enhance the extrapolation ability by improving the Relative Positional Encoding in Transformers, the backbone of most LLMs \cite{su2021roformer,press2021train,luo2022your, vaswani2017attention}. These two kinds of modifications both show substantial improvement in language modelling (LM).
The third category of studies focuses on the recurrent compression of memory for long-range sequence learning \cite{rae2019compressive, peng2023rwkv}.
This methodology effectively learns the comprehensive representation of context, demonstrating notable proficiency in rapid computation and cost-effectiveness during inference. Though the methods above show strong performance in specific tasks and support LLMs with extra-long context windows, i.e. GPT3.5-Turbo-16K, Claude-v1.3-100K and Longchat \cite{longchat2023}, LLMs fail to produce correct answers if related documents are located in the middle of the context, called \textbf{\emph{"lost in the middle"}} \cite{Liu_Lin_Hewitt_Paranjape_Bevilacqua_Petroni_Liang}. It is fatal for Multi-doc QA. However, whether a similar deterioration exists in Chinese LLMs has been unexplored and solutions to this problem have rarely been researched. 


We hypothesise that the scale of attention scores of the beginning context grows large after pre-training and instruction tuning while that of the middle context, whose position is less trained, remains small for a long distance to the current token. This limits the contribution of related information to the answer and results in lower QA accuracy.

To overcome the pitfall, we proposed position-agnostic decompositional training to even up the attention scores over input context. Concretely, we designed a tailored Multi-doc QA task in which positive documents are located at arbitrary positions in contexts among noisy documents. The task presents a significant challenge, compelling the models to extract and summarize information despite the interference of useless ones \cite{ye2022towards}. As human beings routinely solve complex tasks by decomposition to obtain higher quality outcomes \cite{cheng2015break,correa2023humans}, we modified the Multi-doc QA task as a multi-step reasoning task, called \textbf{P}osition-\textbf{A}gnostic \textbf{M}ulti-step QA (\textbf{PAM} QA), combining the Chain-of-Thought (COT, \citeauthor{wei2022chain}) and position-agnostic Multi-doc QA. Trained with explicit extraction of the question and the index of supporting documents before generating answers, models learn to distinguish correct information from noisy ones and attend to them. It also forces attention to the question and supporting indexes 
stronger although the attention scale decays with increasing distance \cite{su2021roformer}.
Empirical results on Multi-doc QA and other benchmarks show that, with only 1/2 or 1/4 context window size, our model improves upon state-of-the-art (SOTA) models by 7.0\% in the top-ranked setting and by 13.7\% in the shuffled setting. Competitive results are shown in other attention-dependent tasks including passage retrieval and summarization. 

\begin{figure*}[tp]
\centering
\includegraphics[width=0.8\textwidth,height=0.34\textwidth]{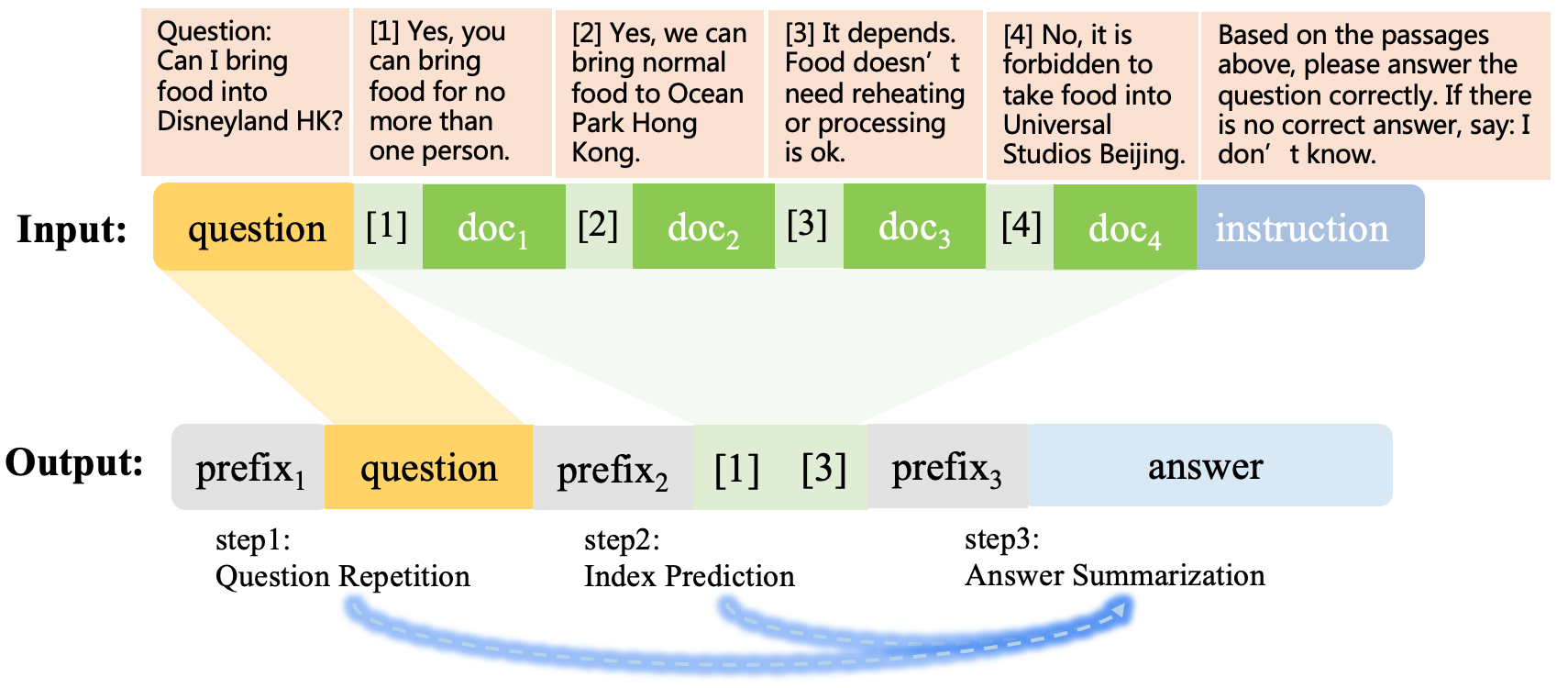}

\caption{The workflow of PAM QA. The blue dashed lines indicate information flows. The desired output of a sample is composed of three parts, corresponding to three steps: Question repetition, index prediction, and answer summarization. [i] refers to the index of the $i$-th document. An input sample is displayed on the top.}
\label{fig_framework}
\end{figure*}


The contribution of this paper is threefold:
\begin{itemize}
    \item This paper proposed a novel task named PAM QA to tackle the "lost in the middle" issue, which is fatal for knowledge-intensive scenarios. To our knowledge, it is the first attempt to solve the problem by training models on special tasks.
    \item We investigate the model's behaviour in-depth, revealing that failing to focus on target information may be the cause of "lost in the middle". 
    \item Comprehensive experiments have shown that the proposed PAM QA is effective in solving the "lost in the middle" problem. Our model surpasses SOTA in Multi-doc QA and other related tasks on renowned Chinese benchmarks. It is non-trivial that the general QA ability of the model is also strong and satisfying. The model is open-sourced to boost future research in the community.
\end{itemize}

\section{Position-Agnostic Multi-step QA}

Multi-doc QA refers to a type of QA task where a model is presented with multiple documents and asked to answer questions correctly. It is difficult for models and humans alike, requiring accurate retrieval, information aggregation and comprehension from noisy candidates while struggling with fading memory. 

In this situation, task decomposition, identifying subproblems and reasoning about them, becomes essential \cite{correa2023humans}. We decomposed the difficult Multi-doc QA to PAM QA. This innovative task comprises three steps, as depicted in Figure 1. 

The entire process of PAM QA unfolds as follows: when receiving a question, a set of candidate documents, and a specific instruction, the model initiates by generating $prefix_1$. It then proceeds to restate the question, predicting the indexes of related evidence after incorporating a connecting phrase, denoted as $prefix_2$. Finally, it formulates an answer by aggregating previous information, following an answer indicator, $prefix_3$.


\subsection{Question repetition}
The first step is question repetition (QR). The questions are placed at the front as a contextual-aware representation \cite{Liu_Lin_Hewitt_Paranjape_Bevilacqua_Petroni_Liang}. The subtask is started with $prefix_1$, "As for the question:" (or expressions with identical meaning) to prompt the model. 

\subsection{Index Prediction}
Supporting evidence not only helps LLMs verify themselves but also aids users in evaluating responses \cite{menick2022teaching}. Remarkable results have been shown in generating quotes and citations \cite{thoppilan2022lamda,menick2022teaching}. We hypothesize that the indicator helps to encode and navigate the attention to corresponding documents.

Accordingly, the second step is index prediction (IP), namely to predict the indexes of the supported documents for the question as an MRC task, beginning with $prefix_2$: "Based on the information numbered". Unlike previous works that predict a verbatim quote extracted from a longer source retrieved, the indexes of corresponding evidence are the targets. For the case in Figure \ref{fig_framework}, the label of this step is "Based on the information numbered [1],[3]". 
Considering the indexes in the second step only count for very few tokens and are hard to emphasize in the sequential cross-entropy loss during training, an MRC task that only asks to predict indexes of correct documents is added as a supplement. 

\subsection{Answer Summarization}
The third step is to generate the final answer after information aggregation. Thanks to the steps above, it can be simplified as an answer summarization (AS) task. The step starts with an indicator like "my answer is" as $prefix_3$.

In line with the proverb \textit{"the palest ink is better than the best memory,"} we teach the model to take notes, turning these annotations into a highway to the relevant knowledge. It can reduce the distraction of extraneous information and make the attention to the question and supporting index stronger because the attention scale decays with increasing distance. 

\section{Training Data Construction}
\label{data construct}
We equipped our model with distinguishing ability through instruction tuning. The training procedure is composed of two stages. We expand the LLM's context window to 8K in the first stage. In the second stage, the model was trained with PAM QA data to solve the attention (or memory) failure called "lost in the middle".

\subsection{Context Window Expansion}

We used about 300k selected data for general supervised finetuning (SFT). The data cover various categories of tasks including QA, MRC, role-playing, writing, coding, translation, brainstorming, math, Language Modeling (LM), and other natural language understanding (NLU) tasks like text classification. The data are packed to 8K window size in a multi-turn conversation style except for the LM task, which calculates the cross-entropy loss on the whole sequence.
 
\subsection{PAM QA}
Data are constructed by formatting inputs and concatenating target outputs of the steps in PAM QA. We first generated Multi-doc QA data and adapted it to PAM QA data. 
 
 First, we filtered out 30K samples of the Fact category with a single answer from DuReader2.0 dataset \cite{he2018dureader} and 20K samples from WebCPM \cite{qin-etal-2023-webcpm}. DuReader2.0 is the largest Chinese MRC dataset collected from Web documents and community QA, containing 200K questions, 420K answers and 1M documents. To ensure the quality of data, we creatively utilize a reward model to score the samples and select the high-quality part of them with thresholds, inspired by \citeauthor{li2023self}. The reward model is trained with 69K human-ranked samples for alignment in general tasks, following \citeauthor{kopf2023openassistant} and \citeauthor{ouyang2022training}. As both datasets only contain positive samples, negative samples are ingeniously generated subsequently.
 
  As collaborative learning is beneficial to RAG \cite{izacard2022few}, we built a search engine with all the documents in the corresponding dataset. For each sample, documents in the whole collection except the positive ones are regarded as negative samples. We retrieved documents from the search engine as negative candidates for a partition comprising 70\% of the data, while we randomly sampled from the original negative candidates for the remaining portion of data. The retrieved negative samples are more relevant to questions and harder to distinguish from the positive ones than random samples. Next, documents are shuffled within each sample in 50\% of the data to prevent positive ones from consistently being at the beginning of contexts. 
 Next, 25K samples were sampled from retrieval benchmarks, T2Rank \cite{xie2023t2ranking} as the relevance MRC, a supplement for task 2. The negative samples are randomly sampled from the hard negative collections and shuffled with positive candidates. The indexes of positive documents are recorded. 
 
The max length of each sample is sampled from 1K to 8K under the uniform distribution. This ensures our model can deal with samples of various input lengths with correct documents located at any position. 
 
To enable the model to recognize situations where the correct document is absent, we generate "Synthetic Unknown" samples, where all documents are negative. The answer for these samples is a constant term indicating "I don't know." This category of data accounts for a proportion of 5\%.

Finally, We sampled some general SFT data, taking a 20\% ratio in this stage to alleviate the catastrophic forgetting \cite{McCloskey_Cohen_1989,rebuffi2017icarl}. The total training samples in this stage summed up to 90K.

\subsection{Training}
We trained our model based on a pre-trained LLM that adapted from LLaMA2, called Ziya2-13B-Base \cite{touvron2023llama,gan2023ziya2, zhang2023fengshenbang}. We trained for 2 epochs on 16 A100 GPUs in both stages with constructed data. The learning rate began with 1e-5 then decayed to 1e-6 with a warmup for the first 0.05\% steps in the first stage. The max learning rate for the second stage was 5e-6. Flash Attention \cite{dao2022flashattention} was utilized to accelerate the training procedure. 
Sampling is turned on for all models during testing in the benchmarks. The hyperparams for testing are listed in Appendix \ref{sec:appendix0}.

\section{Experiments}
In this section, we evaluate the long-context QA abilities of our model and existing representative LLMs. By inspecting the performance, we can verify whether our model overcomes the so-called "lost in the middle" problem \cite{Liu_Lin_Hewitt_Paranjape_Bevilacqua_Petroni_Liang}. 

\subsection{Benchmarks}
We conducted experiments on a long context benchmark, LongBench \cite{bai2023longbench} and  Retrieval-Augmented Generation Benchmark (RGB, \citeauthor{chen2023benchmarking}). LongBench measures various abilities of the testee given long input contexts. Specifically, we tested models on four related tasks in LongBench: Chinese Multi-doc QA, Synthetic tasks, summarization and single-doc QA. We also used the noise robustness testbed in RGB to test the QA ability in short texts, which examines the information extraction ability given a certain ratio of noise documents. 

The synthetic task is a document retrieval task, where given a summary, the goal is to find the corresponding document from a large number of candidates. This task evaluates the information retrieval ability of LLMs in long contexts.
The summarization task gives extremely long meeting records from multiple speakers and asks for a summary. It assesses the model's memory and summarization capabilities.
Single-doc QA is a long-context QA task that is less similar to multi-doc QA. We conduct experiments on this task to test the robustness of the model. 
The context lengths and other statistics of the datasets are listed in Table \ref{longbench_data}. The evaluation scripts were provided by the LongBench official website\footnote{https://github.com/THUDM/LongBench} and RGB official repository.

We also re-constructed the synthetic task to examine whether the models are "lost in the middle". The correct passages are relocated at the 1st, 5th, 10th, 15th and 20th with passages located beyond the 20th removed. The results are in Figure \ref{fig:lost}.

Considering that the documents in the samples of Multi-doc QA tasks are basically sorted by relevance, we shuffled the first 10 candidate documents in each sample to make the real performance exposed, called Multi-doc QA shuffled. 

\begin{table}
\centering
\resizebox{\linewidth}{!}{
\begin{tabular}{lccc}
\hline
\textbf{Datasets} & \textbf{Avg length} & \textbf{Source}& \textbf{Metrics} \\
\hline
Multi-doc. & 15,768& DuReader & Rouge-L \\
Synt. &6,745 & C4 Chinese&Accuracy\\
Summ.& 15,380 & VCSUM& Rouge-L \\
Single-doc & 6,701 & Multifield QA & F1 \\
RGB NR.& 1,105.7 & Self Generated& EM \\ 
\hline
\end{tabular}
}
\caption{The statistics of input lengths of the testing datasets. Multi-doc. is short for Multi-doc QA. Synt. and Summ. represent Synthetic Tasks and Summarization respectively while RGB NR is the abbreviation of RGB noise robustness task. }
\label{longbench_data}
\end{table}

In addition, we conducted a comprehensive human evaluation of model capabilities to see if training on PAM QA harms the general abilities of LLM. The test set contains 200 questions from a wide range of categories. 

\begin{table*}[tp]
\centering
\begin{tabular}{lcccc}
\hline
\textbf{Model} & \textbf{Multi-doc QA} & \textbf{Synthetic Tasks} & \textbf{Summarization} &\textbf{Single-doc QA}\\ \hline
(Baichuan2-Turbo-192K)& 36.8 & 90.0 & 18.4 & 44.7\\
\hline
Longchat-v1.5-7B-32K & 19.5 & 7.6& 9.9 & 29.1 \\
ChatGLM2-6B-32K &37.6& 64.5&\textbf{16.1} & 32.8\\
(ChatGLM3-6B-32K) & 44.8& 94.0& 17.8 & 62.3 \\
\hline
GPT3.5-Turbo-16K & 28.7& 77.5 & 16.0 & \textbf{61.2} \\
Vicuna-v1.5-7B-16K & 19.3&5.0 &15.1 & 43.0\\
\hline
Xgen-7B-8K & 11.0 & 3.5 & 2.2 & 14.8 \\
InternLM-7B-8K & 16.3&0.9 & 12.4 & 33.6 \\
Qwen-14B-Chat & 18.7 &40.0 &13.9& 31.4\\
Our model& \textbf{44.6}& \textbf{98.5} & 15.6 & 34.4\\
\hline
\end{tabular}
\caption{\label{longbench-all}
The results are Rouge-L percentage for Multi-doc QA and Summarization while Synthetic Tasks compute the accuracy (EM scores). Models are separated in lines by context window sizes. ChatGLM3-6B-32K and Baichuan2-Turbo-192K are new models after our work.}
\end{table*}
\subsection{Baselines}
We compared the performance of the most popular LLMs with a long context window. These strong baselines include:
GPT3.5T-Turbo-16K extends the context window to 16K tokens, while both Longchat-v1.5-7B-32K \cite{longchat2023} and ChatGLM2(3)-6B-32K \cite{du2022glm} further push the boundary to 32K tokens. Vicuna-v1.5-7B-16K \cite{zheng2023judging} and Xgen-7B-8K \cite{nijkamp2023long} offer fine-tuned models on user-shared conversations and 8K sequences respectively. Baichuan2-13B-Chat \cite{yang2023baichuan} stands out in few-shot learning with a 4K token window, alongside a larger closed-source variant. Lastly, Qwen-14B-Chat introduces a 14B parameter model with dynamic NTK \cite{dynamic_ntk}, trained on a window size of up to 8K tokens. We refer to retrieval-augmented models as those trained with retrieval-augmented data or paradigms. Baichuan2-13B-Chat and Baichuan2-Turbo-192k are both retrieval-augmented models \cite{yang2023baichuan}.

\section{Results and Discussion}
In this section, we analyze the experimental results of the LLMs and discuss the reason for the findings. An ablation study is also conducted for in-depth attribution. Other details are in the Appendix.

\subsection{Longer window size does not guarantee better performance}
As shown in Table \ref{longbench-all}, our model has a Rouge-L of 44.6\% in the Multi-doc QA task, 7.0\% higher than ChatGLM2-6B-32K, which was the SOTA model. With only 1/4 window size, our model can outperform ChatGLM2-6B-32K at this task. It reveals the strong attention ability of our model since it is an open-book QA task. This Chinese Multi-doc QA dataset does not need to consider all of the contexts, as the correct documents are located at the beginning of contexts.

In the Synthetic Task, namely an abstract retrieval task, our model achieves the highest result with an accuracy of 98.5\%, among models with longer context capabilities. This indicates that the "lost in the middle" issue is almost solved by the proposed method in this paper, as long as the average length is covered. 

As for summarization, ChatGLM2-6B-32K and GPT3.5-Turbo-16K have similar performance with different context window sizes, showing that longer context window sizes do not guarantee better performance. The Rouge-L of our model is only 0.5\% lower than SOTA, without any summarization data in the PAM QA training. As the average length of the task is much longer than 8K tokens, our model with a longer context length will have a promising improvement.

We observe a moderate result in Single-doc QA from our model and find it competitive among 8K models. GPT3.5-Turbo-16K achieves the highest result of 61.2\% F1 score (before ChatGLM3-6B-32K), surpassing the longest model, Baichuan2-Turbo-192K.


\begin{figure}[h!]
    \centering
    \includegraphics[width=.5\textwidth]{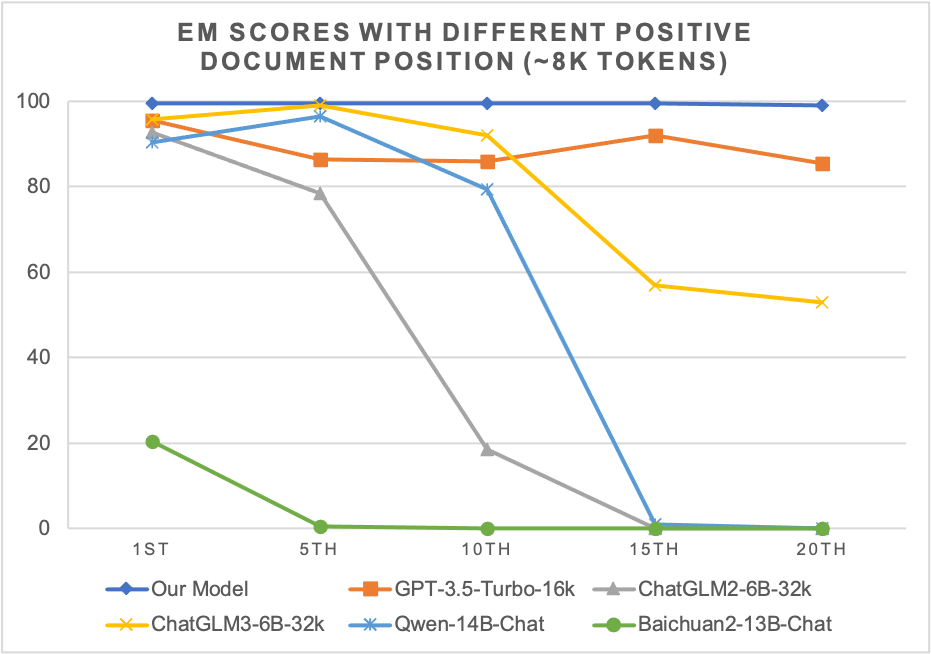}
    \caption{The EM score on Synthetic (passage retrieval) Task from LongBench with correct document inserted to certain position ranging from 1st to 20th.}
    \label{fig:lost}
\end{figure}

\subsection{PAM QA alleviates lost in the middle (and tail) problem}
Experiments on the re-constructed Synthetic (passage retrieval) benchmark display the performance of models on different positive document positions. Concretely, the correct passage for each sample is inserted into the 1st, 5th, 10th, 15th, and 20th locations respectively among the other documents in each experiment. Theoretically, we should see a U-curve described in \cite{Liu_Lin_Hewitt_Paranjape_Bevilacqua_Petroni_Liang}, called "lost in the middle". Results are displayed in Figure \ref{fig:lost}.

However, the figure suggests that most open-source LLMs are lost not only in the middle but also in the tail. A significant decrease is observed when positive documents are placed at the 10th position. Despite the employment of techniques such as Alibi \cite{press2022train} or NTK to expand the context window (i.e. Baichuan2-13B-Chat and Qwen-14B-Chat), models still demonstrate low results. In contrast, our model can survive in different settings of positions, holding a record of 99\%. It reveals the effects of PAM QA training.

\begin{figure*}[ht!]
\centering
\subfigure[ChatGLM2-6B-32K]{
\label{Fig.sub.1}
\includegraphics[width=0.48\textwidth]{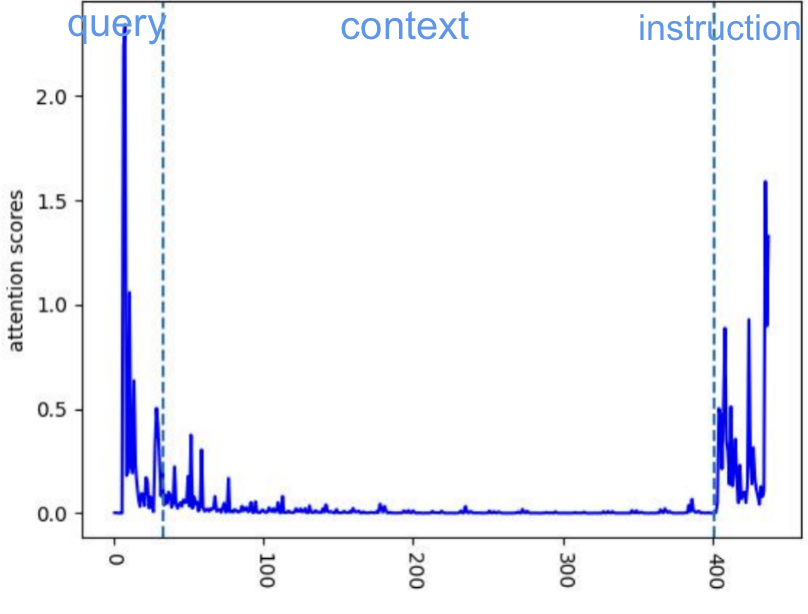}}
\subfigure[Our model]{
\label{Fig.sub.2}
\includegraphics[width=0.48\textwidth]{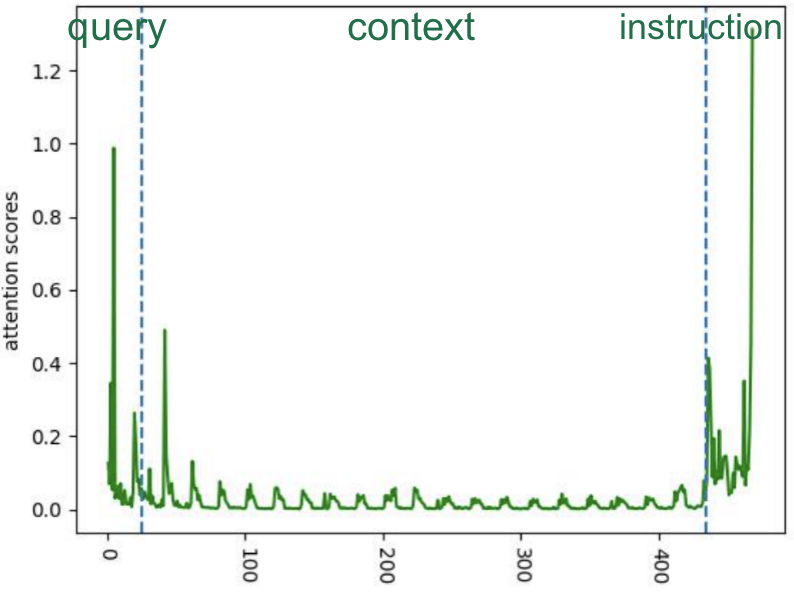}}
\caption{The attention scores over the input tokens in the self-attention procedure within ChatGLM2-6B-32K and our model on a document repeated 20 times. Length differs with tokenizers.}
\label{1}
\end{figure*}

\subsection{Models defeated by shuffled dataset, attention failure being the culprit}

\begin{figure}[ht]
\centering
\includegraphics[width=.5\textwidth]{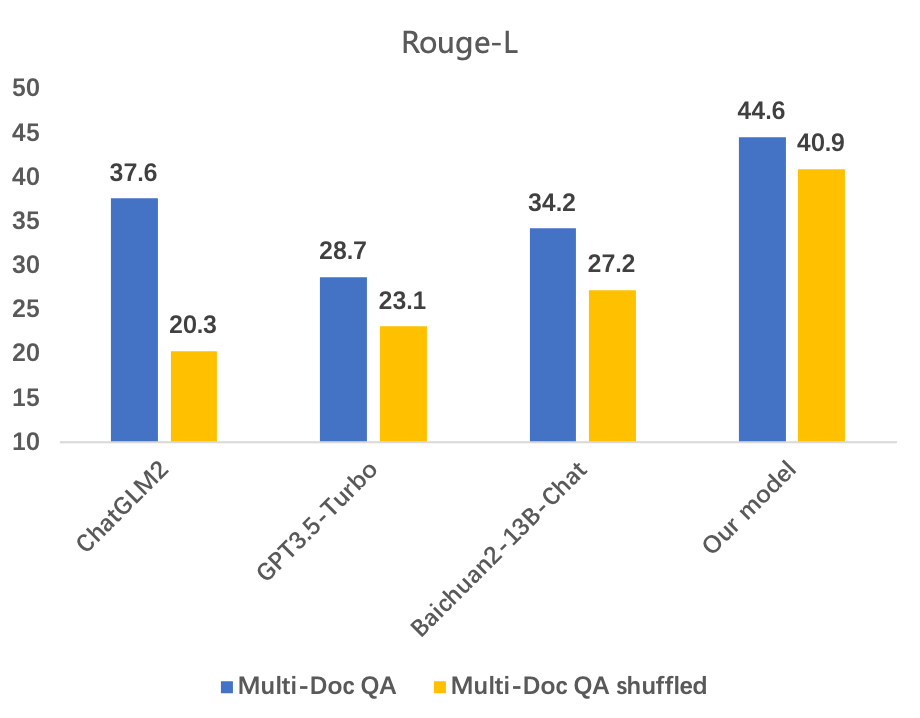}
\caption{Performance on Multi-doc QA before and after shuffling. ChatGLM2 is short for ChatGLM2-6B-32K, GPT3.5-Turbo is short for GPT3.5-Turbo-16K. Scores are in percentage.}
\label{fig_shuffle}
\end{figure}

Figure \ref{fig_shuffle} demonstrates the models' performance on Multi-doc QA before and after shuffling. We can see a sharp decline in all three models except ours. The largest gap reaches 17.3\%, from ChatGLM2-6B-32K. Meanwhile, Baichuan2-13B-Chat also has a 7\% reduction although the shuffled documents are within its context length. Therefore, LLMs without extra long context windows also have difficulty dealing with the challenge. Our model is the most robust model with a 3.7\% decrease. 

To unearth the cause of the decline and examine the attention capabilities of models, we visualize the attention scores of the last layer for the identical input. We repeat a sentence including the correct answer 20 times as the context to find if all of them will be highlighted in the self-attention procedure in models. Attention scores of ChatGLM2-6B-32K and Our model over the input are depicted in Figure \ref{1}. 

We can see the attention scores on documents are fading away in ChatGLM2-6B-32K, as the context after the first 100 tokens is almost neglected. The situation is quite different when it comes to our model. 20 peaks of attention scores are observed (the last one is next to the beginning of instruction), corresponding to the answers in sentences. It reveals that attention to related tokens is the key to the performance gap between models. The models struggle to precisely focus on the correct tokens, paying tremendous attention to the beginning and the ending tokens (where instruction and query are frequently located), which is the culprit of the "lost in the middle" problem.


\subsection{Competitive results observed in short text Multi-doc QA}
\begin{table}
\small
\setlength{\tabcolsep}{3pt} 
    \begin{tabular}{lccccc}
    \hline
    \textbf{Noise Ratio} & \textbf{0} & \textbf{0.2} & \textbf{0.4} & \textbf{0.6} & \textbf{0.8}\\ \hline
    GPT3.5-Turbo& 95.67 & \textbf{94.67}& \textbf{91.00} & \textbf{87.67}& \textbf{70.67}\\ \hline
    ChatGLM2-6B&86.67&82.33&76.67&72.33&54.00\\ 
    (ChatGLM3-6B)&91.67&90.00&89.00&84.67&66.33\\ \hline
    Baichuan2-13B-Chat&93.00&90.33&	89.00	&82.33	&63.33\\
    Qwen-14B-Chat &94.67&92.00&	88.00&	85.30&	69.67\\ 
    Our model&\textbf{96.00}&	90.67&	90.00&	85.50&	67.33\\ \hline
    \end{tabular}
    \caption{\label{RGB}Performance in RGB noise robustness testbed. EM scores are in percentage. ChatGLM3-6B is a new model after our work.}
\end{table}

As reported in Table \ref{RGB}, our model has a competitive performance among open-source models on short-text multi-doc QA although not trained with any short texts. Even compared with the latest popular Chinese LLMs, Qwen-14B-Chat and ChatGLM3-6B-32K, the results of our model are higher under the setting of noise rate in [0,0.4,0.6]. 

\subsection{General ability is preserved with PAM QA Training}
\begin{figure}[h]
    \centering
    \includegraphics[width=0.5\textwidth]{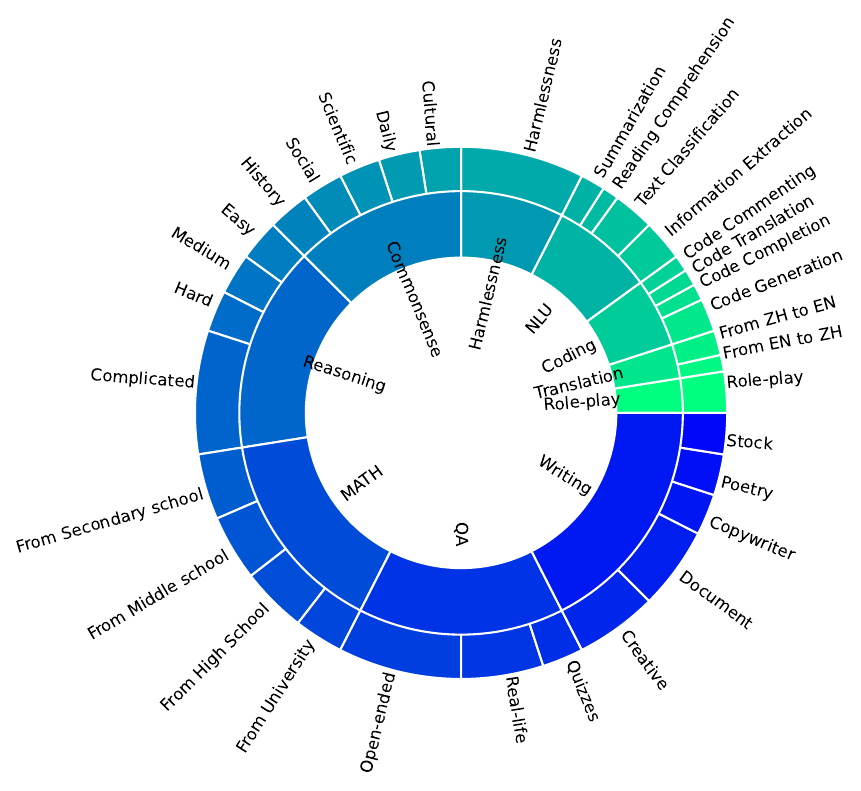}
    \caption{The distribution of tasks in the general ability test.}
    \label{fig:enter-label}
\label{human_tasks}
\end{figure}

\begin{figure}[h]
    \centering
    \includegraphics[width=0.48\textwidth]{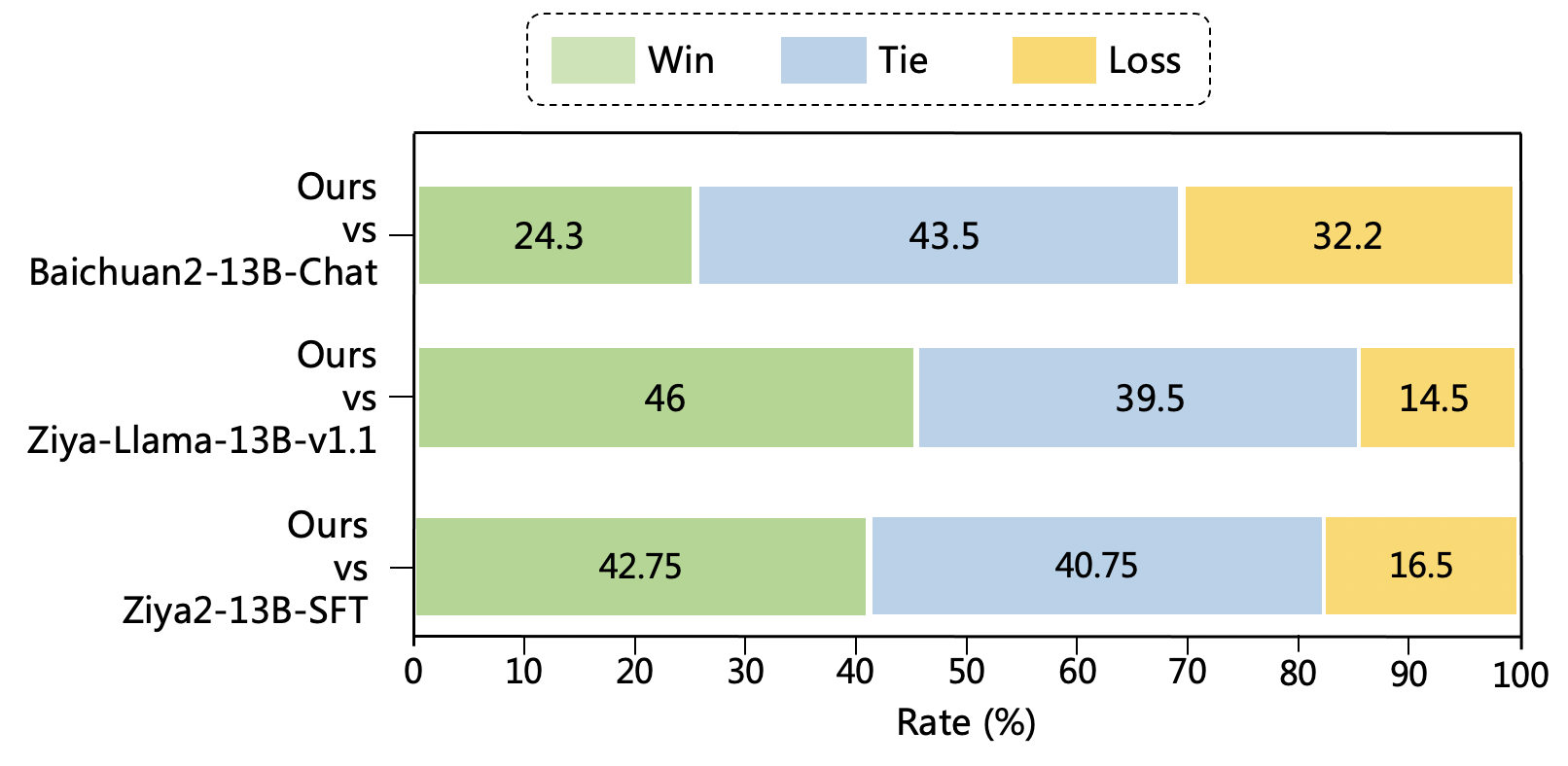}
    \caption{SBS results on general ability evaluation that contains a wide range of tasks. Ziya2-13B-SFT is the full SFT version based on the same pre-trained model. }
    \label{fig:general-table}
\end{figure}

A side-by-side (SBS) comparison was performed by 3 human annotators to check the general ability of our model. General capabilities including commonsense, math, reasoning, QA, writing, harmlessness, etc. are examined in the test, as shown in Figure \ref{human_tasks}. The annotators are asked to choose a better answer among two given answers unless the answers are both bad or the same, as in \cite{zheng2023judging}. Annotators are all master students. They are blind to the models and other information. Results compared with similar size models, Ziya-LLaMa-13B-v1.1\footnote{https://huggingface.co/IDEA-CCNL/Ziya-LLaMA-13B-v1.1} and Baichuan2-13B-Chat respectively are illustrated in Figure \ref{fig:general-table}. We also compare our model with the same base model after full SFT training, Ziya2-13B-SFT.

Figure \ref{fig:general-table} summarizes the human preference between our model and other open-source LLMs. Although our model was trained only with PAM QA data, it performs slightly inferior to Baichuan2-13B-Chat but better than Ziya-LLaMa-13B-v1.1 and Ziya2-13B-SFT significantly. Thus, the general capabilities are maintained after the PAM QA training.

\subsection{Ablation Study}

\paragraph{Each step in PAM QA matters.}
Here we inspect the contribution of each step in PAM QA. The variants are evaluated on Multi-doc QA and Synthetic tasks. Results of this ablation study are listed in Table \ref{ablation}.
\begin{table}[h]
\centering
\small
\begin{tabular}{lcc}
\hline
\textbf{Variants} & \textbf{Multi-doc QA} & \textbf{Synt.} \\
\hline
Our model & \textbf{44.6} & \textbf{98.5} \\
- QR & 38.8 & 98.0 \\ 
- QR - IP & 37.8 & 1.3\\
Only-CWE & 8.7 & 7.5\\
\hline
\end{tabular}
\caption{Synt. is short for Synthetic tasks. Results are in percentage. QR is short for question repetition. IP is short for index prediction. Only-CWE represents the finetuned model only with context window expansion.}
\label{ablation}
\end{table}

Without question repetition, the first step in PAM QA, a 5.8\% decrease can be observed in Multi-doc QA, showing its inevitable contribution to high performance. It strengthens the attention of the question by repeating the question first. Then the model can directly attend to the question in the subsequent steps without going through a long context, reducing the distraction of context when performing self-attention. 

When the index prediction (IP) step is removed, the pronounced decrease in Synthetic tasks emphasizes its importance. It not only teaches LLMs to distinguish between pertinent and irrelevant information but also changes the model's prior behaviour (i.e., seeking information from the beginning and the end of context). Meanwhile, it streamlines the process by allowing models to concentrate on relevant abstracted information, instead of repeatedly scanning extensive input tokens. A performance drop in multi-doc QA also shows the contribution of IP. Since the scale of the attention scores decays as the distance grows \cite{su2021roformer}, models with rotary position embeddings (RoPE) struggle to remember the remote tokens without training. With the former two steps, the question and the potentially correct evidence are listed just a few tokens ahead. This reduces the probability of forgetting questions and context by decreasing the distance.

An enormous gap between the results of Our model and that without QR and IP, indicates the substantial improvement from PAM QA training.
We visualize the attention scores when predicting the first token and discover the generated questions and indexes are highlighted, shown in Figure \ref{fig:heatmap} in Appendix \ref{sec:appendix2}.

Compared with Only-CWE, the variant model without QR and IP steps also gains 29.1\% improvement, which shows the effect of position-agnostic and challenging negative candidates. By transforming the Multi-doc QA into PAM QA, the same data can boost the performance by 6.8\% in Multi-doc QA, and 97.2\% in Synthetic Tasks, which reveals the strength of task decomposition training.

\paragraph{Necessity of Training.}
To investigate whether training (fine-tuning) is necessary, we performed multi-step COT prompting in the style of "first predict the indexes of relevant documents" and "according to the information, the final answer is" on different models. We removed the question repetition step in COT for better performance. Results are in Table \ref{tab:cot}.
\begin{table}[ht]
    \centering
    \small
    \begin{tabular}{lcccc}
    \hline
        Model &	Strategy & Multi-doc. &Synt.  \\ \hline
        GPT3.5-Turbo-16K & w/o COT & 28.7	& 77.0 \\
        GPT3.5-Turbo-16K &w/ COT&	28.3&	63.9 \\ \hline
Yi-34B-Chat&	w/o COT&	14.9	&35.3\\
Yi-34B-Chat&	w/ COT&	2.5&	58.6\\ 
Yi-34B-Reader (Ours) & w/o COT &\textbf{45.1} & 50.4\\ \hline
         Ziya2-13B-SFT&	w/o COT&	11.0&	6.3\\
Ziya2-13B-SFT&	w/ COT&	1.1	&2.1\\
Ziya2-Reader (Ours) & w/o COT & 44.6 &\textbf{98.5} \\
\hline
    \end{tabular}
    \caption{Comparison of results from different strategies: models with (w/), without (w/o) multi-step COT inference and with PAM training (Ours). }
    \label{tab:cot}
\end{table}

As demonstrated by the superior results of our models over the multi-step COT inference, training is essential to optimize performance. Especially in Multi-doc QA, LLMs tend to produce an answer with more hallucination after predicting a list of indexes of related documents without fine-tuning. We found models with multi-step COT struggle to handle complex instructions and maintain long-context memory.

\begin{table*}[th]
    \centering
    \begin{tabular}{c|ccc}
    \hline
        Model & Multi-doc QA & Synthetic Task & Summarization \\ \hline
        Yi-34B-Chat&	14.9&	35.3&	13.8 \\
         Yi-34B-Reader (Ours) &	\textbf{45.1}	&50.4&	14.2 \\
        Ziya2-13B-SFT&	11.0&	6.3	&12.6\\ 
        Ziya2-13B-Reader (Ours)&	44.6&	\textbf{98.5}&	\textbf{15.6}\\\hline
    \end{tabular}
    \caption{Comparison of models trained with PAM and official full SFT models based on the same pre-trained models. Ziya2-13B-SFT is the model trained on the same pre-trained model, Ziya2-13B-Base.}
    \label{tab:generalization}
\end{table*}

\paragraph{Generalizability of Method.}
To illustrate the generalization of the approach, we also performed identical training on another Chinese pre-trained model, Yi-34B-Base{\footnote{\url{https://huggingface.co/01-ai/Yi-34B}}, a top 34B pre-trained model in LLM Benchmarks. Table \ref{tab:generalization} lists the comparison of the model we trained (Yi-34B-Reader) and the official instruction tuning version, Yi-34B-Chat\footnote{\url{https://huggingface.co/01-ai/Yi-34B-Chat}}.

Results show our method can be generalized to other LLMs. However, it is harder to change the behaviour of Yi-34B-Base using only 100K PAM QA data compared to the 13B model since it was pre-trained rather (maybe over) sufficiently. It results in lower results in the synthetic task than the 13B Ziya2-Reader.

 \section{Related Works}
\subsection{Retrieval-Augmented Language Models}
Retrieval-Augmented Language Models (RALMs) mark notable progress in NLP by merging the capabilities of expansive LMs with the precision and intricacy offered by external knowledge sources. \cite{guu2020retrieval,lewis2020retrieval,izacard2022few}.
These models use a retriever to search through a large body of evidence, like Wikipedia, to find specific documents related to the user's query. Afterwards, a reader component is utilized to carefully examine these documents and generate a response. This two-step process guarantees both relevance and depth in the produced answers. Recent research efforts have concentrated on enhancing the performance of the retriever \cite{karpukhin2020dense,sachan2023questions} or the reader\cite{izacard2020leveraging,cheng-etal-2021-unitedqa}, training the system end-to-end \cite{lewis2020retrieval,sachan2021endtoend}, and integrating the retrieval systems with black-box large language models \cite{shi2023replug,yu2023generate,trivedi-etal-2023-interleaving} 

\subsection{RALMs Adapted to Long and Noisy Context}
Recent research emphasizes the influence of contextual length and the position of related context on the performance of LLMs \cite{krishna2023longeval,bai2023longbench,Liu_Lin_Hewitt_Paranjape_Bevilacqua_Petroni_Liang}. The research closely aligned with ours is the study by \citep{yoran2023making}, training RALMs to disregard irrelevant contexts. A homothetic COT-like training approach was proposed to solve math and coding problems, emitting intermediate computation steps into a "scratchpad" \cite{nye2021work}. However, they overlooked long context scenarios, specifically the "lost in the middle" issue, a key consideration in our work.

An earlier work that considered multi-doc modelling in training is proposed by \citeauthor{caciularu2023peek}. After splitting long context into pieces and generating QA pairs based on picked salient ones, they asked models to predict the masked salient sentences and answers, given other pieces and the generated questions. Significant improvement in multi-hop QA benchmarks after fine-tuning with the training set is reported at the expensive cost of pre-training. However, there is no training set in most benchmarks nowadays and it fails to perform diverse tasks in the zero-shot setting.

\section{Conclusion}

In this paper, we assume that the widely recognized "lost in the middle" phenomenon may caused by weak attention to target information. We discover popular Chinese LLMs are "lost" both in the middle and tail. A novel approach is proposed to address the deficiency in LLMs by training models with Posistion-Agnostic Multi-step (PAM) QA. Experimental results show the superiority and effectiveness of our method, surpassing SOTA LLMs in Multi-doc QA and passage retrieval significantly, with only 1/4 context window size. By shuffling the candidate documents in open benchmarks, degraded performance is observed in all models, among which our model is the most robust one. The ablation study also reveals the significant effect of PAM QA and the positive contribution of its components. Our study also finds that LMs with extremely long context windows do not ensure better performance on Multi-doc QA and passage retrieval tasks. We hope our study provides profound insight into the "lost in the middle" problem and sheds light on developing more intelligent LLMs.

\section*{Limitations}
Our work covers the important "lost in the middle" issue and experiments with Chinese Benchmarks on popular Chinese and English LLMs with long context capability. The improvements in tested tasks do not imply similar upgrades in all aspects, like math and reasoning. The constructed PAM QA data were used after or during SFT, with the effect in pre-training and RLHF period unexplored.

The data construction method is mainly based on multi-doc QA and shows substantial gains in related tasks. Improvements in other long-context tasks are not as impressive as multi-doc QA and synthetic tasks since other abilities are more required than discriminating and focusing. Those situations are not considered in this paper. 


The proposed approach is language-independent and could be applied to datasets of other languages. The core of the method lies in constructing samples with challenging related negative documents, diverse positions of positive samples and multi-step reasoning answers. Therefore, the method can potentially alleviate the "lost in the middle" issue in other languages.

\bibliography{anthology,custom}
\bibliographystyle{acl_natbib}

\appendix

\section{Hyper-parameters}
\label{sec:appendix0}
We used the following settings in the inference phrase in the LongBench test:
$do\_sample=True$, $top_p=0.85$, $temperature=0.8$, $repetition\_penalty=1.0$, $early\_stopping=True$.\\
As for the RGB noise robustness testbed, the temperature was set to 0.2.


\section{Evaluation of Index Prediction}
We apply human annotation to 140 samples in Multi-doc QA to investigate the precision and recall of the index prediction (IP) step. The macro precision is 43.0\% and the recall is 70.9\%. They result from more indexes predicted than ground truths. The LLMs may have the reflection ability to check the indexes and finally produce correct answers. Therefore, the IP step benefits the answer generation even in low precision. 

\section{Demonstration of PAM QA}

\label{sec:appendix1}

A case of Multi-doc QA and the answer produced by Our model and ChatGPT3.5 are shown in Table \ref{case}.

\begin{table*}[hbp]
\centering
\begin{tabularx}{\textwidth}{X}
\toprule
Chinese: \\
\midrule
\includegraphics[width=0.98\textwidth]{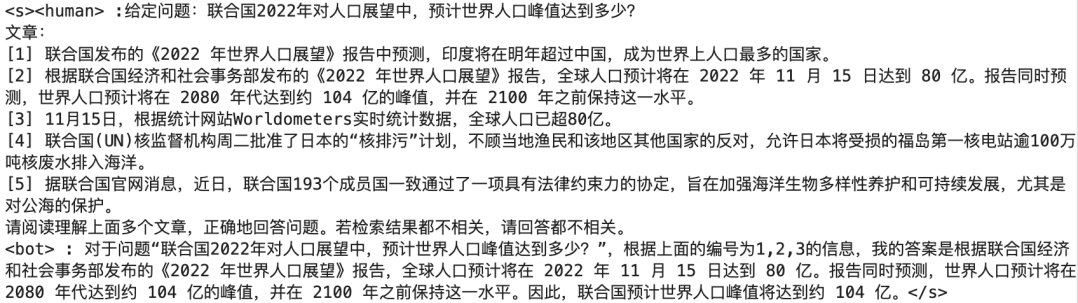} \\
\midrule
English: \\
\midrule
<s><human>: Given question: What is the expected peak world population in the UN's 2022 population outlook? \newline
Essays: \newline
[1] The World Population Prospects 2022 report released by the United Nations predicts that India will surpass China as the world's most populous country next year.\newline
[2] According to the World Population Prospects 2022 report published by the United Nations Department of Economic and Social Affairs, the global population is expected to reach 8 billion by 15 November 2022. The report also predicts that the world's population is expected to peak at about 10.4 billion in the 2080s and remain at that level until 2100.\newline
[3] On November 15, according to real-time statistics from the statistics website Worldometers, the global population has exceeded 8 billion.\newline
[4] The United Nations (UN) nuclear watchdog on Tuesday approved Japan's plan to "dismantle nuclear wastewater," allowing Japan to discharge more than 1 million tons of nuclear wastewater from the damaged Fukushima Daiichi nuclear power plant into the ocean, despite opposition from local fishermen and other countries in the region.\newline
[5] According to the official website of the United Nations, the 193 member states of the United Nations recently unanimously adopted a legally binding agreement aimed at strengthening the conservation and sustainable development of marine biodiversity, especially the protection of the high seas.\newline
Please read and understand many of the passages above and answer the questions correctly. If the search results are not relevant, please answer that they are not relevant.\\
\midrule
<bot> : In response to the question "What is the expected peak world population in the United Nations 2022 population outlook?" Based on the information numbered 1,2,3 above, my answer is that according to the World Population Prospects 2022 report published by the United Nations Department of Economic and Social Affairs, the global population is expected to reach 8 billion on November 15, 2022. The report also predicts that the world's population is expected to peak at about 10.4 billion in the 2080s and remain at that level until 2100. As a result, the United Nations expects the world's population to peak at about 10.4 billion.</s>\\ 
\midrule
ChatGPT3.5: The expected peak world population in the UN's 2022 population outlook is not explicitly mentioned in the provided essay passages.\\
\bottomrule
\end{tabularx}
\caption{A Multi-doc QA sample and the output of Our model, ChatGPT3.5 in the last two cells respectively. The answer of our model starts after the <bot> :. Special tokens are removed when testing ChatGPT3.5. We can see that Our model produces a correct answer while ChatGPT3.5 fails to answer the question.}
\label{case}
\end{table*}

\section{Visualization of attention scores on tokens in answer summarization step}
The heatmap of attention scores of each token when generating the final answer is displayed in Figure \ref{fig:heatmap}. 
\begin{figure*}[ht]
    \centering
    \includegraphics[width=0.9\textwidth]{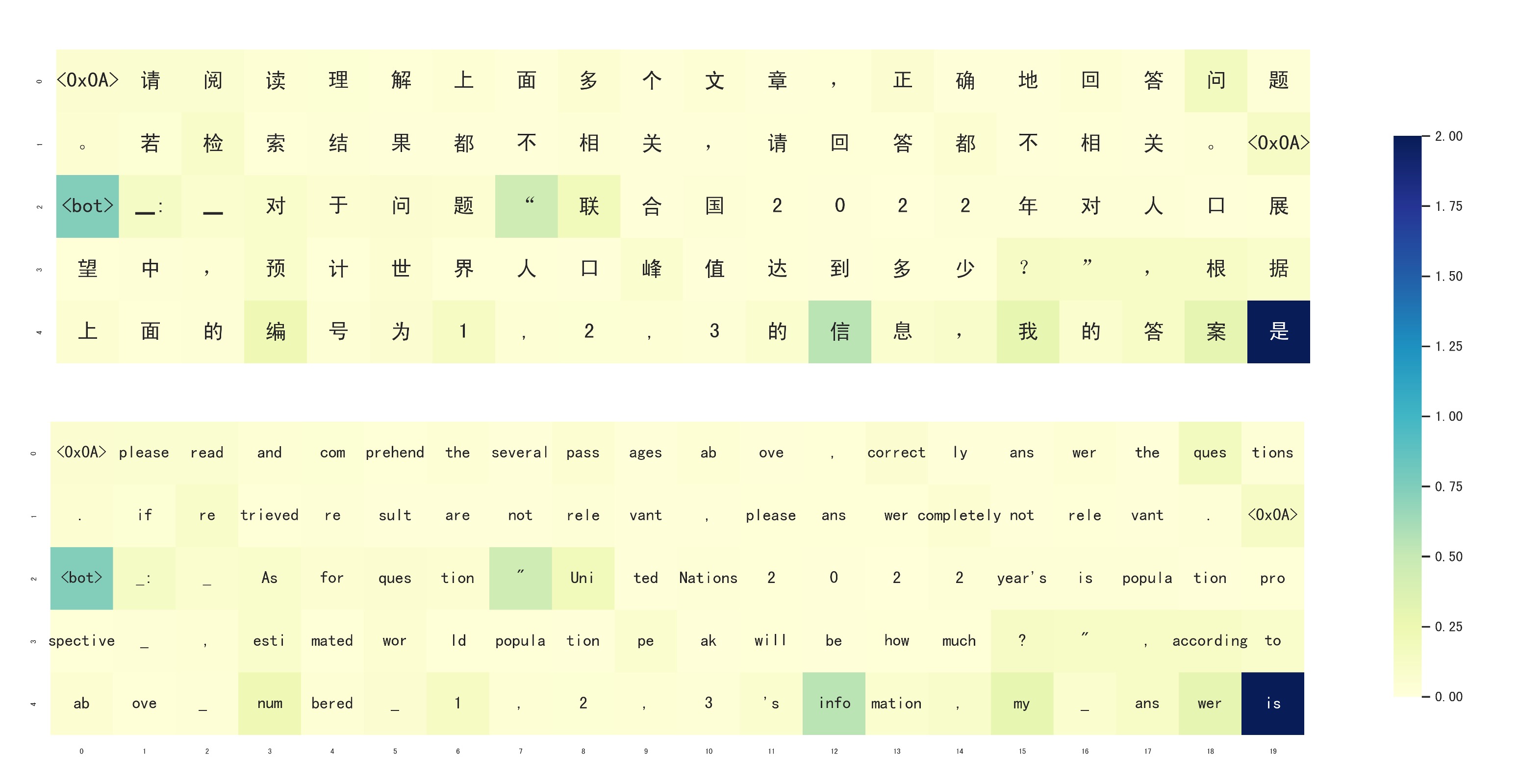}
    \caption{The heatmap of attention scores on the last 100 tokens when our model begins to generate an answer. The darker represents the higher score. The English version is translated from the Chinese heatmap token by token. The first token of the generated question and the first predicted index are attended to, indicating the utility of the question repetition and index prediction in PAM QA.}
    \label{fig:heatmap}
\end{figure*}
\label{sec:appendix2}

\section{Annotation System}
The screenshot of the annotation system in SBS general ability evaluation is shown in Figure \ref{fig:sbs}.
\begin{figure*}
    \centering
    \includegraphics[width=0.9\textwidth]{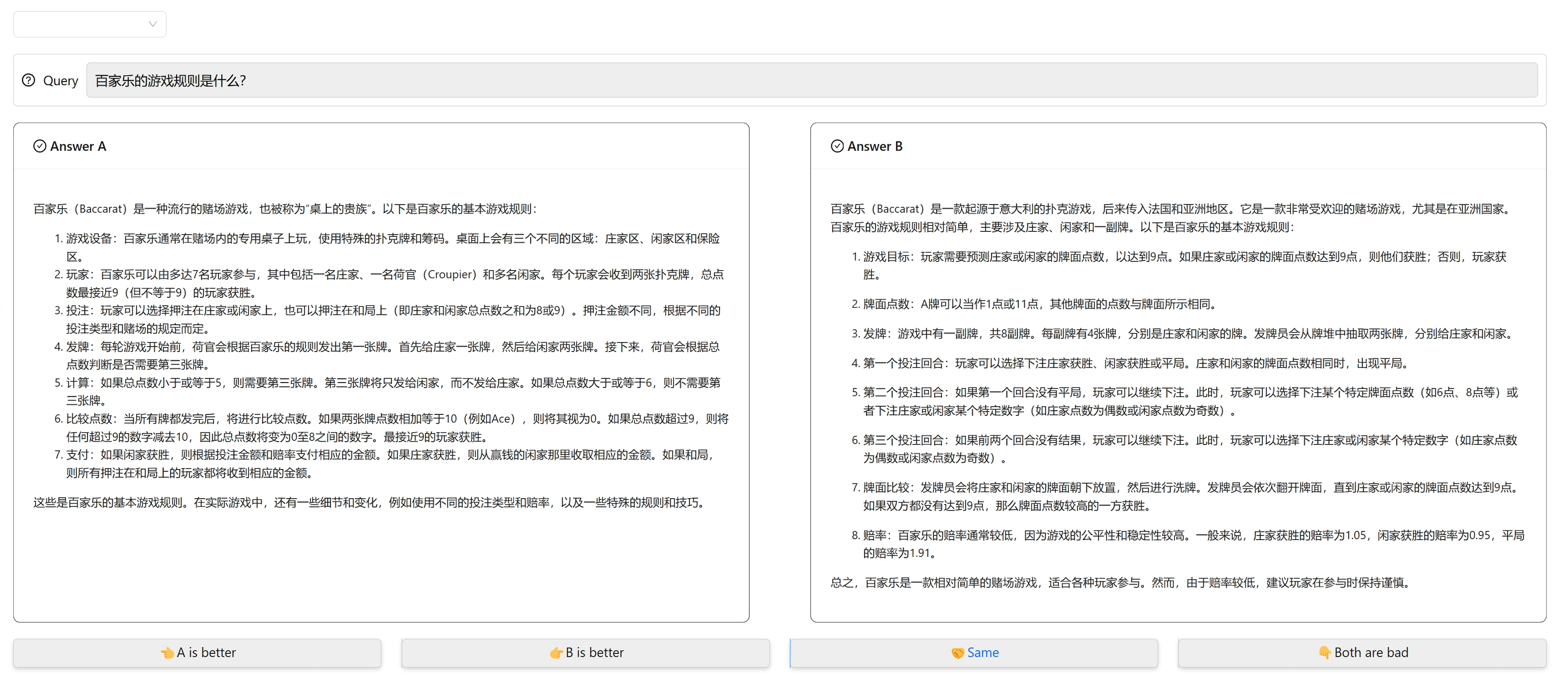}
    \caption{Interface of SBS evaluation system. Users can only rank answers without access to the names of models. }
\end{figure*}
\label{fig:sbs}
\end{document}